\pdfoutput=1
\documentclass[11pt]{article}

\usepackage[T1]{fontenc}
\usepackage[utf8]{inputenc}
\usepackage{lmodern}
\usepackage[margin=1in]{geometry}
\usepackage[english]{babel}
\usepackage{amsmath}
\usepackage{amssymb}
\usepackage{graphicx}
\usepackage{float}
\usepackage{xcolor}
\usepackage{cite}
\usepackage[colorlinks=true,linkcolor=blue!50!black,citecolor=blue!50!black,urlcolor=blue!50!black]{hyperref}
\usepackage{xurl}

\title{3D Neural Rendering for Automated Ultrasonic Laser Scans}

\author{\parbox{0.92\textwidth}{\centering
  Thomas Pierson\textsuperscript{1},
  Matthew Mitchell\textsuperscript{2},
  Kira Perna\textsuperscript{3},
  Anand Iyer\textsuperscript{4},
  Eric Flynn\textsuperscript{4},
  and Adam Wachtor\textsuperscript{4}\\[1.5ex]
  {\small
  \textsuperscript{1}Carnegie Mellon University\\
  \textsuperscript{2}North Carolina State University\\
  \textsuperscript{3}Penn State University\\
  \textsuperscript{4}Los Alamos National Laboratory\\[1ex]
  Corresponding author: \href{mailto:tmp@lanl.gov}{tmp@lanl.gov}}%
}}

\date{LA-UR-26-28755}

\begin{document}

\maketitle

\begin{abstract}
Laser doppler vibrometer (LDV) scans can be used to detect defects and cracks in manufactured parts before they are deployed. This work presents a method for reconstructing the steady-state surface wavefield of a three-dimensional part by combining automated robotic manipulation with neural rendering. A six-degree-of-freedom (6-DOF) robotic manipulator (UFactory xArm6) with an embedded piezoelectric transducer presented an acoustically excited test article to a stationary LDV at a sequence of known orientations, yielding a set of complex-valued 2D scans with poses derived analytically from the arm kinematics rather than from image-based pose estimation. These scans were used to train a novel complex-nerfacto model, a modified Neural Radiance Field (NeRF) pipeline that replaces the standard three-channel RGB output head with a two-channel output. Complex-nerfacto represents the real and imaginary velocity components of a wavefield, preserving the complex measurement without quantization. Reconstructions of a welded aluminum cylinder and a welded T-bracket were evaluated against measured scans using Z-normalized cross-correlation (ZNCC). The complex-valued pipeline achieved a mean ZNCC of 0.60 at held-out viewing angles, compared with 0.28 for a baseline RGB nerfacto model trained on rendered images of the same data. The resulting surface wavefield reconstructions supported defect detection and characterization on complex components. We identify geometry reconstruction from sparse LDV scans as the remaining obstacle to practical deployment.
\end{abstract}


\section{Introduction}

Defects and damage in mechanical components can often be detected through the analysis of elastic wave propagation. In acoustic non-destructive evaluation (NDE), a structure is excited using an ultrasonic transducer to generate elastic waves. Structures with relatively thin walls can be modeled using Rayleigh--Lamb wave theory \cite{bib1}. This theory describes the steady-state response of a structure as the superposition of multiple symmetric (S) and anti-symmetric (A) wave modes. In NDE, the first wave modes (A0 and S0) are most often used to isolate defects \cite{bib2}. Because these modes propagate at different geometry and material-dependent speeds, disparities in propagation velocity tend to concentrate in sudden transitions regions. This property makes surface wave velocity analysis a useful framework for characterizing cracks, defects, or delamination.

Laser Doppler Vibrometry (LDV) is a scanning technique used to measure the local surface velocity of a vibrating structure \cite{bib3,bib22}. Two common LDV-based approaches exist for acquiring waveform data. The first uses a pulsed laser to thermally excite the structure's surface, followed by a scanning laser that captures the resulting transient wave modes \cite{bib4}. The second technique, used in this work, is Acoustic Steady-State Excitation Spatial Spectroscopy (ASSESS). ASSESS uses a transducer that provides continuous steady-state excitation while an LDV records the velocity response across the structure. The LDV beam is steered by a pair of galvanometer mirrors that raster-scan the measurement area, which allows relatively fast imaging of the steady-state response over a spatial grid \cite{bib5}. This response is analyzed in the wavenumber domain and can be used to infer local thickness variations and detect embedded defects \cite{bib6}.

The majority of prior ASSESS literature has focused on flat-plate specimens. Extending this technique to the more geometrically complex, three-dimensionally shaped parts common in manufacturing requires a different approach to data collection and visualization. This paper describes an approach toward reconstructing the full three-dimensional (3D) surface wavefield of a part rather than a single 2D projection of it.

Another challenge in guided wave spectroscopy is the reconstruction of a dense wavefield from a limited number of measurements. This has previously been addressed using compressed sensing and sparse wavenumber recovery \cite{bib20,bib21}. Those methods are formulated primarily for planar geometries and known dispersion models, not generalized to 3D parts or to partial views of a non-planar surfaces.

Neural Radiance Fields (NeRFs) offer a promising route to this reconstruction. NeRFs generate synthetic views of a 3D scene from a sparse set of 2D images by learning a continuous volumetric function that maps spatial position and viewing direction to color and density \cite{bib7}. Subsequent work has improved NeRF sampling, training speed, and practicality \cite{bib15,bib16,bib14}. Beyond conventional photographic rendering, NeRFs have been adapted to other imaging mechanisms, including hyperspectral imaging \cite{bib8} and acoustic fields \cite{bib9}, showing that the architecture can generalize to multi-channel, non-RGB data. This generalizability is necessary for ASSESS data, which is inherently quadrature components of the steady-state response, rather than three-channel RGB. A NeRF is also a useful tool for interpolating between views, which matters here because the measured surface velocity component changes as the LDV viewing angle changes.

To collect the data necessary for 3D reconstruction, a method for acquiring scans from many viewing angles was developed. In this paper, a six-degree-of-freedom (6-DOF) robotic manipulator was used to reorient the test article between scans. Serial robotic arms are increasingly used to automate NDE inspection in manufacturing settings \cite{bib10,bib23}. Robotic automation also offers a particular advantage for glovebox applications \cite{bib11}: rather than moving a sensitive and expensive LDV unit into a harsh environment, the manipulator can reposition the test article while the LDV remains stationary and protected.

In this study, transducers are embedded inside the robotic gripper, and the arm presents an acoustically excited part at a sequence of controlled orientations to a stationary LDV scanner, capturing an ASSESS scan at each pose. This dataset of 2D scans is then pre-processed and used to train a modified NeRF architecture that reconstructs a 3D representation of the part's surface wavefield response.

Key contributions of this work include:

\begin{itemize}
\item Instrumentation of a 6-DOF robotic arm to apply steady-state acoustic excitation to a test object across a range of controlled orientations.
\item A method for deriving camera poses for 3D reconstruction directly from serial-manipulator kinematics, rather than from conventional image-based camera-pose estimation.
\item Modification of the nerfacto architecture to train directly on real and imaginary velocity channels in place of RGB color data, which avoids information loss from rendering complex velocity data to 8-bit images.
\item Reconstruction of the 3D surface wavefield of two welded test articles, with novel-view renderings evaluated quantitatively against measured scans.
\end{itemize}

\section{Methods}

\subsection{Problem Formulation}\label{S_problem}

Although ASSESS has been used extensively to scan structures in two dimensions, no existing method describes a process for collecting scans across many viewing angles and combining them into a single 3D reconstruction. The first challenge in building such a reconstruction is understanding the structure of the underlying LDV data.

When a structure is excited at a steady-state frequency $f$, the vibrational velocity response at a single point can be written as the superposition of two quadrature components at that same frequency:
\begin{equation}\label{EQ_vt}
v(t) = A \cos(2\pi f t) + B \sin(2 \pi f t).
\end{equation}
Equivalently, the steady-state response at each scan point can be extracted from a sampled record $v(t_n)$ by evaluating the discrete Fourier component at the excitation frequency:
\begin{equation}\label{EQ_complex}
c = \frac{1}{N} \sum_{n=0}^{N-1} v(t_n)\, \exp(-j 2 \pi f t_n),
\end{equation}
where $j$ is the imaginary unit and $N$ is the number of samples acquired at each point, spanning an integer number of excitation periods. This demodulation is phase-locked to the excitation reference so that all scan points, and all scan poses, share a common phase origin.

The LDV records a velocity-proportional voltage at each scan point, which is converted to surface velocity using the decoder calibration constant; all values reported hereafter are in m/s. For a full 2D scan, the resulting complex-valued matrix $c_{x,y} = A_{x,y} + jB_{x,y}$ contains all relevant steady-state response information, stored as complex 64-bit data (two 32-bit floating-point values per pixel) by an NI-USB 6363 data acquisition unit. The real and imaginary components, $A_{x,y}$ and $B_{x,y}$, can be interpreted intuitively as two snapshots of the steady-state response separated by a $90^{\circ}$ temporal phase shift.

A single-head LDV is sensitive to the component of surface velocity projected onto the laser beam direction. For a flat plate scanned at normal incidence this is well approximated by the out-of-plane velocity, but for non-planar geometry and oblique incidence angles the measured quantity is $\hat{d} \cdot \mathbf{v}$, where $\hat{d}$ is the beam direction. The measured field is therefore view-dependent by construction. This quality of a surface velocity field makes the NeRF methodology, in which the emitted quantity $c(x, d)$ is also a function of viewing direction, a useful framework for 3D reconstruction.

\subsection{Visualization}\label{S_viz}

This complex-valued scan data can be visualized in several complementary ways. The real and imaginary components can be plotted separately as individual color maps, or combined to construct complex gradient maps, which are useful for highlighting local wave discontinuities associated with damage. Alternatively, the phase relationship between the real and imaginary components can be used to generate a temporal animation of the wave propagating across the structure by iterating through $t$ values in Equation \eqref{EQ_vt}.

One useful representation is the local complex gradient magnitude, computed via the Wirtinger ($\bar\partial$) derivative of the complex wavefield $c = A + jB$:
\begin{equation}\label{EQ_wirtinger}
\bar\partial c = \frac{1}{2}\left(\frac{\partial c}{\partial x} + j\frac{\partial c}{\partial y}\right).
\end{equation}
The magnitude of this quantity, $g = |\bar\partial c|$, highlights regions of high velocity differential in the wavefield, typically associated with defect boundaries. This quantity has a physical meaning: a local plane wave $c \sim \exp(j\mathbf{k}\cdot\mathbf{x})$ is used to find the gradient magnitude, which scales as $|\mathbf{k}||c|$. In this case, $g$ acts as a proxy for local wavenumber. Because local wavenumber depends on thickness and stiffness, $g$ concentrates at the material and geometric transitions described by Rayleigh--Lamb theory.

\begin{figure}[H]
  \centering
  \includegraphics[width=1\textwidth]{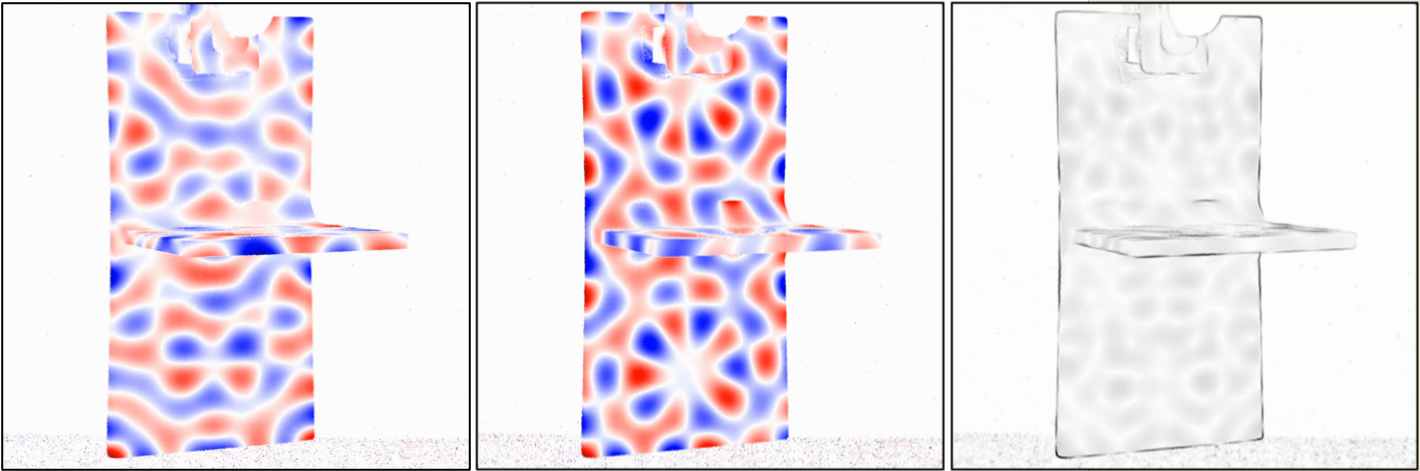}
  \caption{Real, imaginary, and gradient representations of the complex-valued LDV scan data. An externally invisible crack can be seen in the complex gradient map.}\label{F_viz}
\end{figure}

Each of these visualization modalities allows an inspector to identify cracks or other defects using different, complementary features of the same underlying dataset.

\subsection{Neural Radiance Fields}

Prior to the introduction of NeRFs, two primary methods existed for reconstructing a 3D scene from a collection of 2D images. Structure from Motion (SfM) uses feature detection and matching both to estimate camera poses and to build a sparse 3D point cloud of the scene \cite{bib12}. Next, Multi-View Stereo (MVS) used a method of assuming known camera poses to produce a dense point cloud by tracking neighboring views. Subsequently, neural-network-based methods were developed that encode the scene as a learned latent representation rather than an explicit point cloud \cite{bib13}, called Neural Implicit Fields (NIFs).

A NeRF model represents a scene as a continuous radiance field, and a neural network is trained to map a 3D position and viewing direction to a color and a volume density. To render a pixel, a camera ray
\begin{equation}\label{EQ_ray}
r(t) = o + td
\end{equation}
is cast through the scene, where $o \in \mathbb{R}^3$ is the ray origin (camera center), $d \in \mathbb{R}^3$ is the normalized ray direction, and $t$ is distance along the ray. At each sampled point $x \in \mathbb{R}^3$ along the ray, the network predicts a density $\sigma(x)$ and a color $c(x, d)$. The ray is then discretized into $N$ sample points $\{r(t_i)\}$ with step sizes $\delta_i = t_{i+1} - t_i$, and the final rendered pixel value is computed via volume rendering along the ray:
\begin{equation}\label{EQ_volint}
\hat{C}(r) = \sum_{i=1}^{N} T_i \, \alpha_i \, c_i, \qquad
\alpha_i = 1 - \exp(-\sigma_i \delta_i),
\end{equation}
where $T_i = \exp\left(-\sum_{k<i}\sigma_k \delta_k\right)$ is the accumulated transmittance up to sample $i$. This integration is fully differentiable, so the network is trained using gradient descent across the whole ray. Training minimizes an $L_2$ photometric loss between the rendered color $\hat{C}(r)$ and the ground-truth color $C(r)$, summed over all training rays (i.e., pixels from the training images):
\begin{equation}\label{EQ_loss}
\mathcal{L} = \sum_{r} \left\| \hat{C}(r) - C(r) \right\|_2^2.
\end{equation}

A well-known limitation of neural networks is their difficulty learning high-frequency variation. In images, this corresponds to sharp changes in color or depth between neighboring pixels. NeRFs address this using positional encoding, which remaps low-dimensional spatial and directional inputs into a higher-dimensional space using sinusoids of increasing frequency, making high-frequency variation easier for the network to represent. For example, each coordinate of the input position $x = (x, y, z)$ is encoded as:
\begin{equation}\label{EQ_posenc}
\gamma(p) = \left(\sin(2^0\pi p),\ \cos(2^0\pi p),\ \ldots,\ \sin(2^{L-1}\pi p),\ \cos(2^{L-1}\pi p)\right),
\end{equation}
with an analogous encoding applied to the viewing direction $d = (\theta, \phi)$. The baseline model used in this work is nerfacto, the default method in the Nerfstudio framework \cite{bib14}, which combines a multiresolution hash encoding \cite{bib16} with the proposal sampling and scene contraction strategies of mip-NeRF \cite{bib15}.

\begin{figure}[H]
  \centering
  \includegraphics[width=1\textwidth]{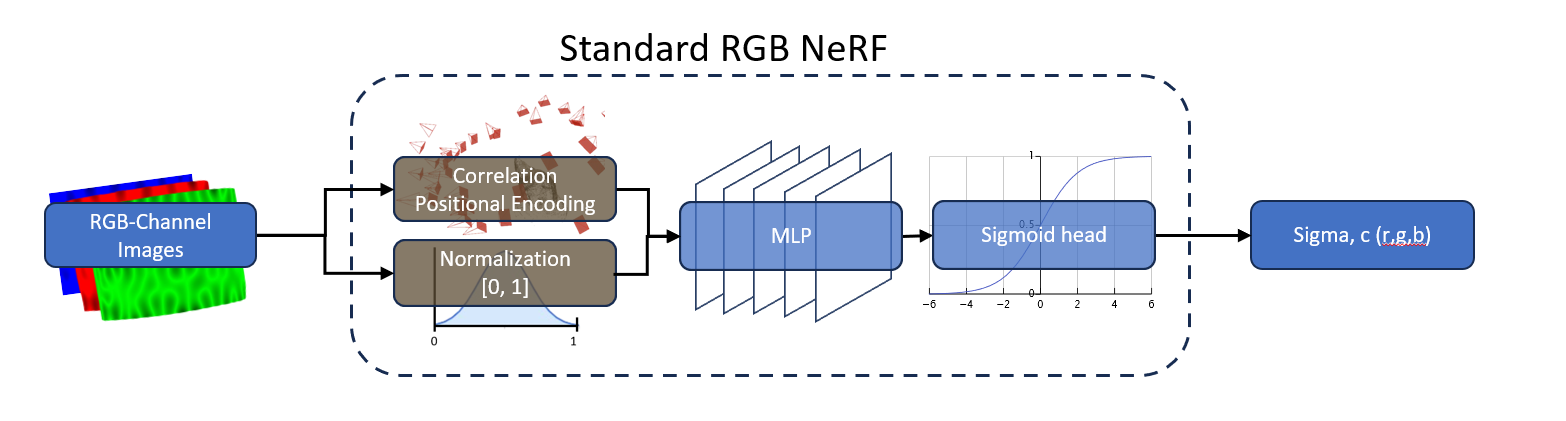}
  \caption{Baseline nerfacto model, which uses a photometric loss and a sigmoid output activation.}\label{F_nerfacto}
\end{figure}

\subsection{Adapting NeRF for Static LDV Scans}

The standard NeRF formulation assumes an RGB scene captured by a moving camera around a static object. Our setup inverts both of these assumptions: the LDV is stationary, and the object itself is reoriented by the robotic arm between scans. In addition, LDV data is single-channel complex-valued (complex64) rather than three-channel, 8-bit RGB. A conceptual distinction is also worth emphasizing: because the LDV measures surface velocity rather than reflected light, the field being learned is not a photometric radiance field but a surface velocity field. The ``color'' learned by the network corresponds to a physical vibration amplitude rather than an appearance property. Volume rendering is retained as the differentiable image-formation mechanism, but the quantity being reconstructed exists only on the illuminated surface of the part; no interior wavefield is measured or recovered.

To reconcile these differences with the Nerfstudio pipeline, we made two categories of changes: pose reconstruction and data formatting.

\paragraph{Pose reconstruction.} Because the object rotates while the LDV remains fixed, we reconstruct an equivalent ``virtual'' camera trajectory using the known rotation angle of the test article at each scan (relative to its initial orientation) together with the fixed LDV standoff distance. This produces a synthetic camera pose for each scan, consistent with the standard NeRF assumption of a camera orbiting a static object, even though physically it is the object that moved. To isolate the rotating part from the stationary background, each LDV scan is masked so that only the test article and the robotic gripper (both of which rotate together) remain in frame.

\paragraph{Data formatting.} Each LDV scan is split into two float32 matrices corresponding to the real and imaginary velocity components. These are passed to a modified Nerfstudio data parser that accepts two-channel float32 inputs in place of standard 8-bit, three-channel RGB images, reducing the input tensor's channel dimension from three to two. Because the training data are not conventional images, Nerfstudio's built-in interactive viewer could not be used directly; instead, novel views were rendered offline at a sequence of specified angles after training to generate the final reconstructions.

\begin{figure}[H]
  \centering
  \includegraphics[width=1\textwidth]{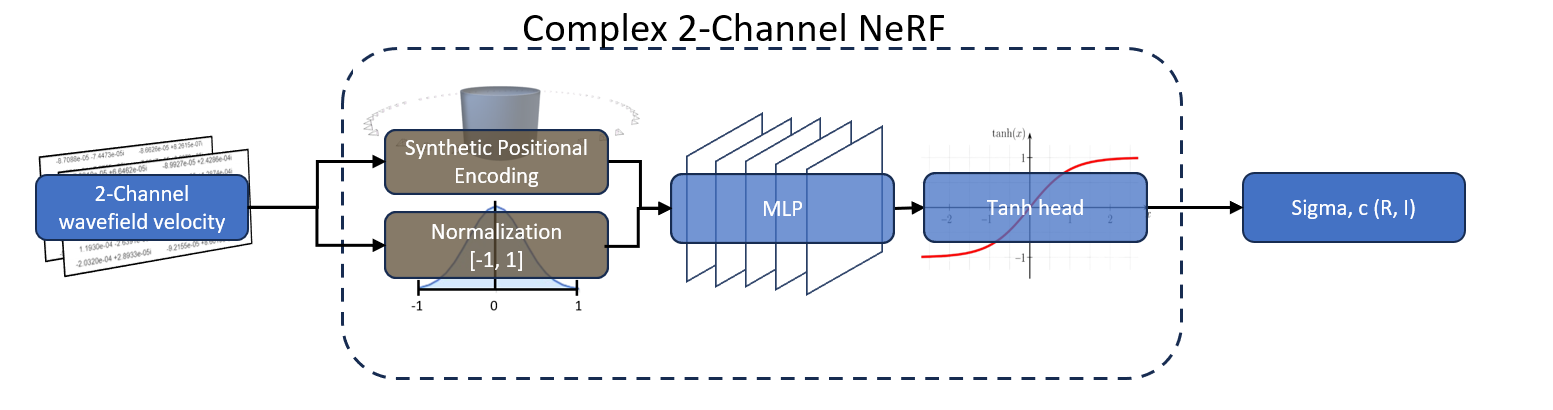}
  \caption{The repurposed complex-nerfacto model for LDV scans, which trains on rendered velocity loss with a tanh output activation.}\label{F_cnerf}
\end{figure}

\subsection{Custom Complex-Nerfacto for LDV Data}

The goal of this adaptation was to determine whether a NeRF-style architecture could reconstruct a 3D surface vibration wavefield from a set of 2D LDV scans. The difficulty is that the nerfacto pipeline is built around ordinary photographs, with three RGB channels bounded to $[0, 255]$ and camera intrinsics estimated from imagery. For our ASSESS data, each scan is a complex-valued array with real and imaginary velocity components that are signed and small in magnitude (approximately $\pm 0.026$ m/s in our dataset). Feeding this data into a standard NeRF pipeline as rendered images would discard half the information (the imaginary channel) and introduce quantization and clamping loss in the remainder.

To preserve the structure of the data rather than distort it to fit NeRF, we developed a custom Nerfstudio method, complex-nerfacto, by subclassing three core pipeline components:

\begin{itemize}
\item \textit{Data parser:} reads camera geometry from a \texttt{transforms.json} file but references \texttt{.mat} files in place of images.
\item \textit{Dataset loader:} loads each complex-valued array and returns it as a two-channel tensor formed by stacking the real and imaginary components, rather than a three-channel RGB tensor.
\item \textit{Field (network head):} replaces the standard three-channel sigmoid output head with a two-channel tanh head. Sigmoid outputs are restricted to $[0, 1]$ and therefore cannot represent data distributed about zero, whereas tanh outputs span $[-1, 1]$, matching the signed normalized velocity data.
\end{itemize}

The LDV sensor velocity magnitudes of around $10^{-2}$ m/s were too small to be used for training with the tanh output head. This data was normalized, similar to the Nerfstudio's standard normalization, except that our data was normalized to $[-1, 1]$ using a global dataset scale. The bounds for normalization were created using the largest-magnitude velocity value across the full scan set, which ensured symmetric normalization bounds (This kept value consistency between NeRF outputs and scan outputs, which standard RGB-nerfacto struggled with).

These modifications repurposed the photometric radiance field model as a surface velocity field model. Ray marching and transmittance accumulation were left unchanged from the standard NeRF architecture.

\section{Experiments}

\subsection{Experimental Setup}

The experimental setup consists of four integrated hardware subsystems. A scanning laser Doppler vibrometer (LDV) is steered by a motorized mirror gimbal, which traces the laser scan path along a part. A 6-DOF robotic manipulator holds and reorients arbitrary components, and an embedded transducer inside of the robot hand excites the held part.

\begin{figure}[H]
  \centering
  \includegraphics[width=0.65\textwidth]{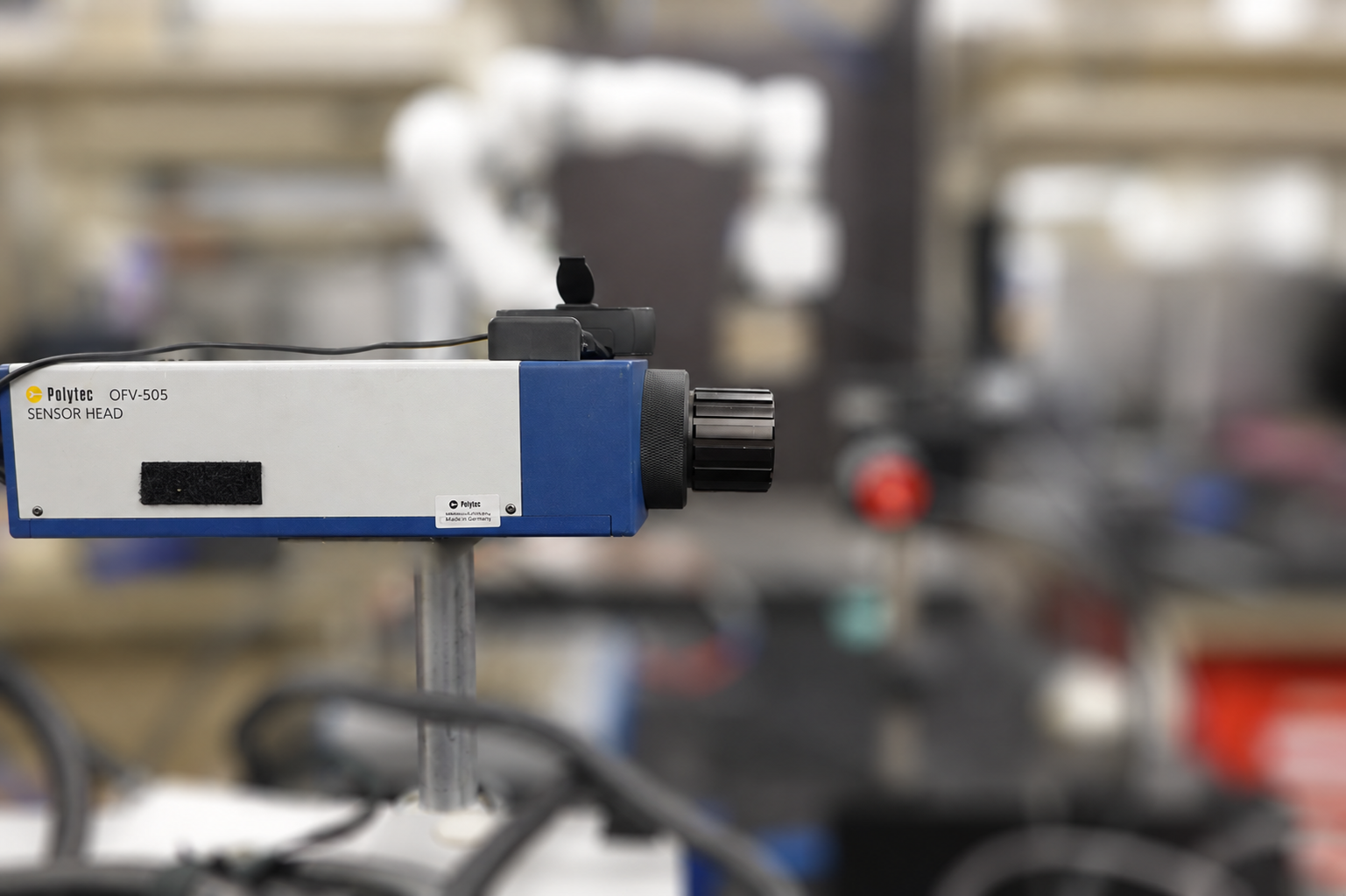}
  \caption{Laser doppler vibrometer sensor head.}
  \label{F_ldv}
\end{figure}

\paragraph{Laser Doppler Vibrometer.} Surface velocity was measured with a Polytec VibroFlex Q-Tech LDV (Fig.~\ref{F_ldv}). The sensor head measures oscillatory surface velocity in the direction of the beam at the contact point. The LDV uses the Doppler frequency shift of reflected laser light, where surface velocity can be inferred using the shift in reflected frequency of the laser. The unit supports a measurement bandwidth of 0 to 32 MHz. Our components were generally excited in the 50--90 kHz range, so no aliasing considerations were necessary for scan data.

\paragraph{Mirror Manipulation.} The LDV beam was rastered over an arbitrary scan window using a motor-actuated mirror steering gimbal. The beam is projected across a 2-axis plane under closed-loop control for repeatable positioning of each scan point. The beam pivots about a fixed point at a fixed standoff, so the resulting sampling geometry was approximated by a pinhole perspective centered at the gimbal pivot. This was important for determining camera intrinsics for 3D reconstruction.

\begin{figure}[H]
  \centering
  \includegraphics[width=0.65\textwidth]{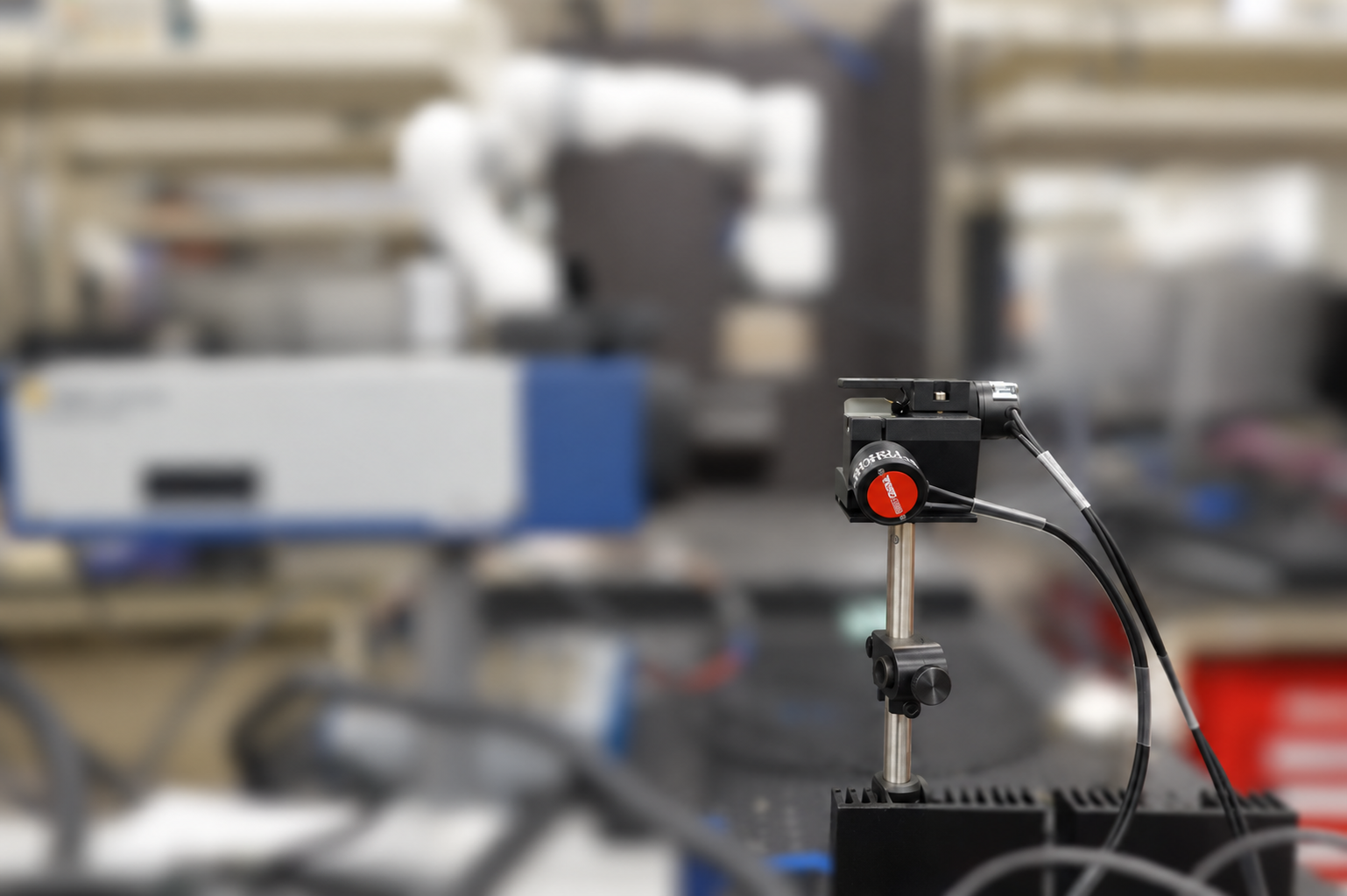}
  \caption{Laser steering gimbal with galvanometric mirrors.}
  \label{F_gimball}
\end{figure}

\paragraph{Robotic Manipulator.} A UFactory xArm6 serial manipulator with six degrees of freedom was used to pick up and reorient each test article between scans. This robot arm enabled fully automated data collection with accurate angular labels. This also simplifies the scanning process, since the LDV itself does not move while generating full-field scans. A UFactory four-bar-linkage end effector was mounted to the arm to grip each test article.

\begin{figure}[H]
  \centering
  \includegraphics[width=0.65\textwidth]{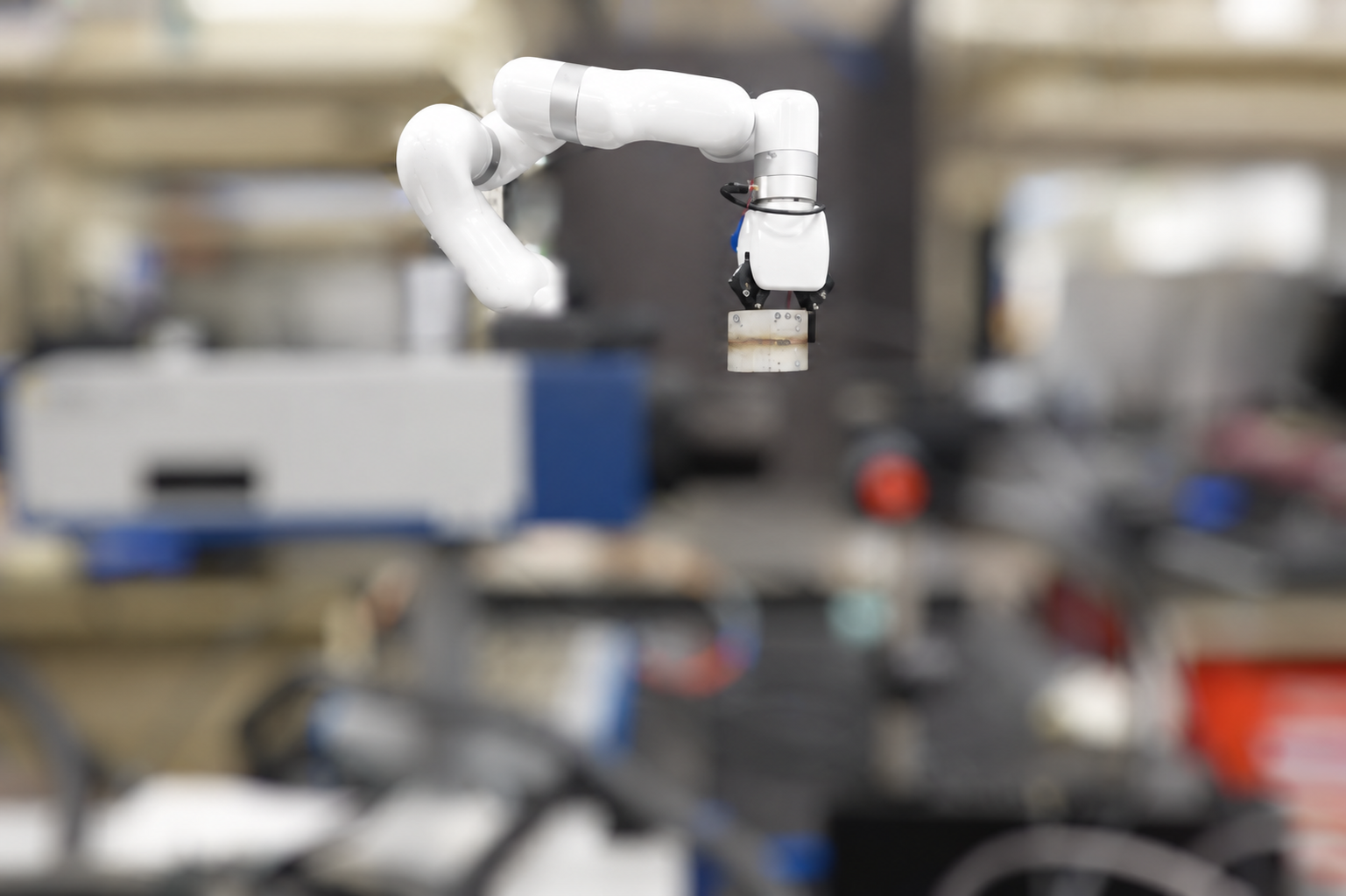}
  \caption{Ufactory 6DOF serial manipulator, gripper, and experimental component.}
  \label{F_arm}
\end{figure}

\paragraph{Acoustic Excitation.} Each test article was driven into steady-state Lamb-wave vibration using a piezoelectric lead-zirconate-titanate (PZT) transducer mounted inside the robotic gripper. The transducer used a wrap-around electrode so that both signal and ground leads could be connected on the non-contact face of the wafer. The transducer was acoustically decoupled from the arm using a silicone isolation pad. The excitation signal was a sinusoidal waveform generated by an NI-DAQ signal generator, amplified, and delivered continuously through the transducer during each scan \cite{bib26}.

\subsection{Test Articles}

Three test articles were used to evaluate the system across different geometries and defect types, summarized in Table~\ref{t1}.

\begin{table}[htbp]
\centering
\caption{Test articles and associated defects.}\label{t1}
\begin{tabular}{|p{0.42\textwidth}|p{0.42\textwidth}|} \hline
\textbf{Test Article} & \textbf{Defect} \\ \hline
Flawtech T-bracket & Externally-invisible crack along weld line \\ \hline
Welded thin-wall aluminum cylinder & Through-crack along weld line; weld splattering \\ \hline
\end{tabular}
\end{table}

\begin{figure}[H]
  \centering
  \includegraphics[width=0.7\textwidth]{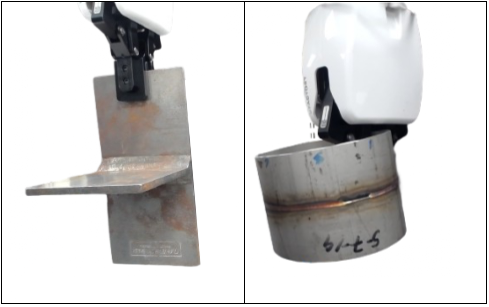}
  \caption{Flawtech T-shaped welded bracket (left), welded cylinder (right).}\label{F_parts}
\end{figure}

\subsection{Automated Data Acquisition Procedure}

Each test article was mounted in the robotic gripper such that the center of rotation of the arm coincided with the top of the gripper, which was in turn aligned with the center of the LDV scan window. This alignment ensures that the physical rotation axis of the part matches the assumed rotation axis used later for pose reconstruction.

A scan-automation script allows the user to specify the number of scan poses to be captured around the equator of the part. The script then commands the xArm6 to step through evenly spaced rotation angles about this axis, pausing at each pose to trigger an LDV scan of the excited part. At each pose, the LDV records a 2D raster grid of complex-valued velocity following the formulation described in Section~\ref{S_problem}. This produces a dataset of paired complex-velocity scans and labeled with rotation angles suitable for further 3D reconstruction. All camera poses exist on a single orbital path around the object, so the reconstructed field is constrained to this visible orbit. The top and bottom faces of components were not observed, and not necessary for defect detection.

\subsection{Pose Reconstruction and Intrinsics Validation}

Because the object rotates instead of the camera in our setup, camera poses cannot be recovered using conventional structure-from-motion techniques, which assume a moving camera and static scene. Instead, an artificial camera trajectory was constructed analytically. The part rotates counterclockwise about a fixed axis, so we created an equivalent stationary-object formulation by rotating a virtual camera clockwise along a circular trajectory about the same axis, at a radius equal to the LDV standoff distance. Our scanning loop converts each recorded arm rotation angle into a corresponding $4\times4$ quadrature transform.

\subsection{Preprocessing: Background Masking}

Each complex-valued scan contains velocity information both from the test article and from empty background in the LDV scan window. In order to isolate the rotating test article, we used a salient-object segmentation model (rembg, based on a U$^2$-Net architecture) \cite{bib18,bib19} to mask out scan backgrounds. In order to use this, the magnitude of the complex velocity at each pixel was rendered as a grayscale image, then segmented, and used as a black-and-white mask. These black and white masks were used to filter the complex data files by zeroing all background pixels. The resulting mask was then applied directly to the original complex matrix by zeroing all background pixels. This was the most consistent masking method we found to ensure no velocity information belonging to the test article was discarded in the process.

\subsection{Complex-Nerfacto Reconstruction Experiment}

To validate the full pipeline, the custom complex-nerfacto model was trained on 59 masked scans of the vibrating aluminum cylinder, each paired with an artificial camera pose derived from the corresponding xArm6 rotation angle. The model was trained for 20,000 iterations using the two-channel (real/imaginary) tanh output head previously described.

After training, arbitrary views were generated in a full orbit sweep around the reconstructed part at the same radius as the physical scans. These novel views were then stitched together to create full-field animations of the experimental part. The reconstructed wavefields pattern rotated smoothly, and remained spatially coherent across the full orbit, even at interpolated angles. This continuity indicated that the network learned a latent 3D representation of the surface wavefield rather than simple interpolation. This is most visible on the T-bracket scans, which also maintain part geometry at synthetic view angles.

\section{Results}

Novel views extracted from the trained models show a consistent advantage for the complex-valued pipeline. Both models preserved the overall shape and continuity of the wavefield pattern, but the custom complex-nerfacto renderings were more numerically faithful to the measured velocity data and contained fewer visual artifacts than those of the baseline RGB nerfacto model. This difference is attributable to the complex-nerfacto pipeline training on unquantized real and imaginary channels, whereas the RGB baseline trains on data that has been discretized in both value and channel count.

\begin{figure}[H]
  \centering
  \includegraphics[width=0.8\textwidth]{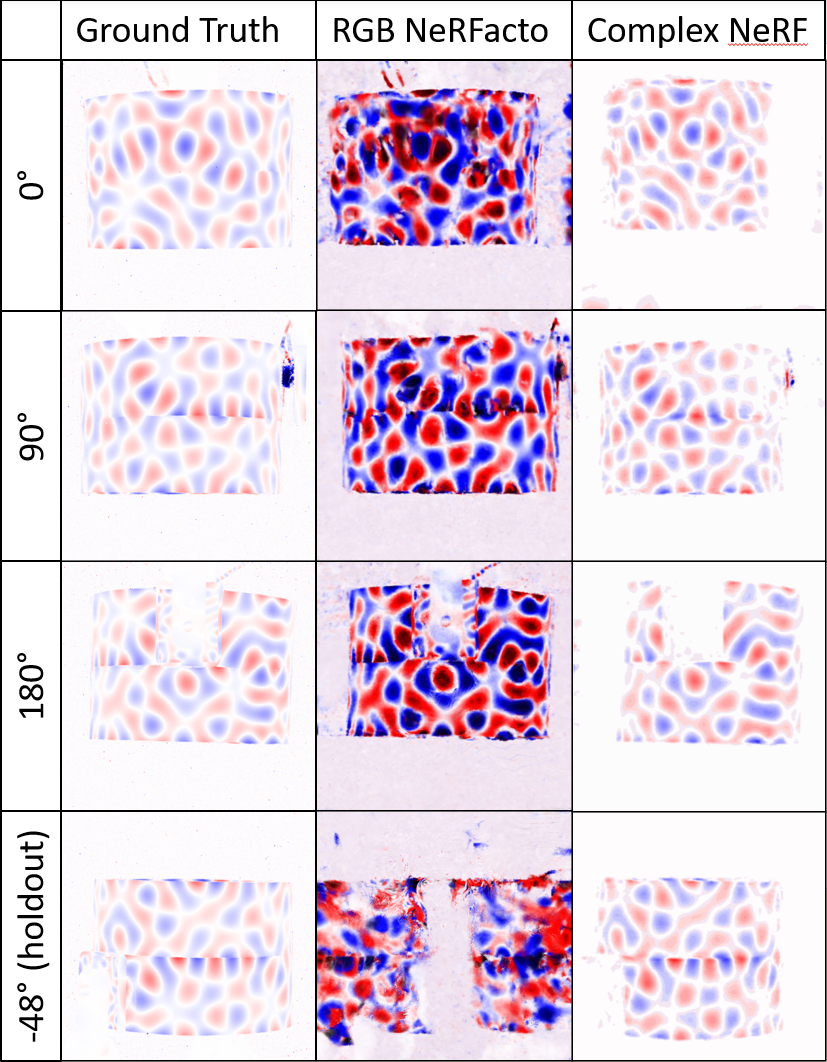}
  \caption{Visualizations of the real velocity component of the nerfacto and complex-nerfacto renderings compared to ground truth for the cylinder.}\label{F_res_cyl}
\end{figure}

These differences are most pronounced at the held-out viewing angles, which were excluded from the training set and therefore best indicate how well each model generalizes across the reconstructed geometry. At the $-48^{\circ}$ holdout pose, complex-nerfacto reproduced the shape and form of the wavefield noticeably better than the RGB model. It also better preserved the smaller-magnitude features of the wavefield and was less prone to the clouding and floating artifacts visible in the nerfacto renderings.

\begin{figure}[H]
  \centering
  \includegraphics[width=0.8\textwidth]{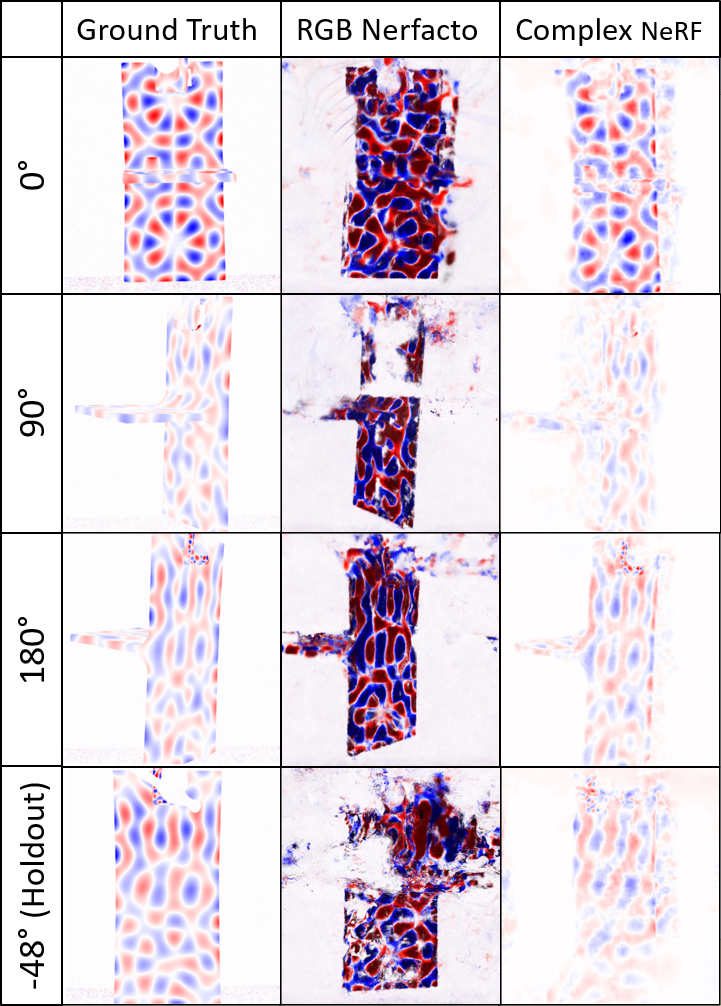}
  \caption{Visualizations of the real velocity component of the NeRF renderings compared to ground truth for the T-shaped bracket.}\label{F_res_flaw}
\end{figure}

To quantify reconstruction fidelity, rendered views at each measured angle were compared against the corresponding ground-truth scan using the Z-normalized cross-correlation (ZNCC) score. ZNCC is a metric that shows the empirical similarity between two images, scored from $-1$ to $1$, where two identical images would show a score of 1, a perfect negative image would be scored as $-1$, and uncorrelated images would score 0. For these results, the ZNCC was calculated for rendered images from the NeRF models at angles where a ground-truth image existed, and these were scored. ZNCC was used as opposed to a metric like mean squared error (MSE), because MSE is not normalized to data scaling. When comparing LDV scan images with very small voltage values to RGB images scaled from 0 to 255, this normalization from 0 to 1 is important for a meaningful comparison between methods.

\begin{equation}\label{EQ_zncc}
\text{ZNCC} = \frac{1}{MN}\sum_{x=0}^{M-1}\sum_{y=0}^{N-1}
\frac{\big(I(x,y) - \bar{I}\big)\big(T(x,y) - \bar{T}\big)}{\sigma_I \, \sigma_T}.
\end{equation}

\begin{figure}[H]
  \centering
  \includegraphics[width=0.7\textwidth]{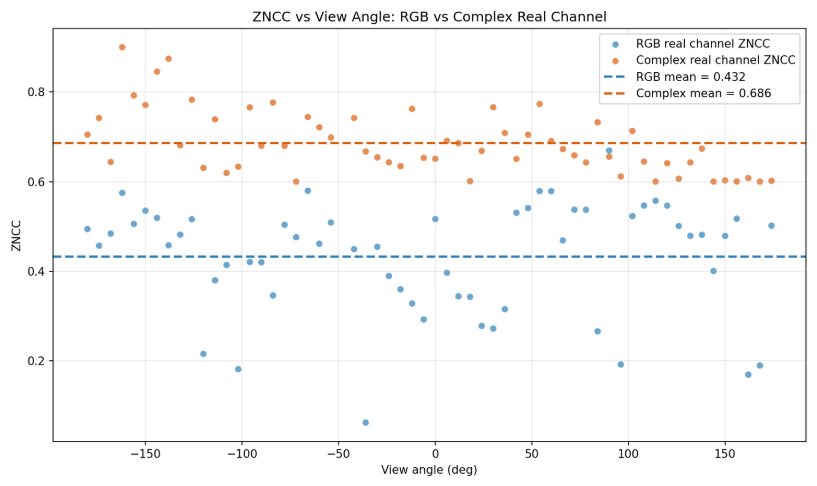}
  \caption{ZNCC relative to ground truth for the RGB and complex NeRF models at each scan angle for the cylinder.}\label{F_zncccyl}
\end{figure}

\begin{figure}[H]
  \centering
  \includegraphics[width=0.7\textwidth]{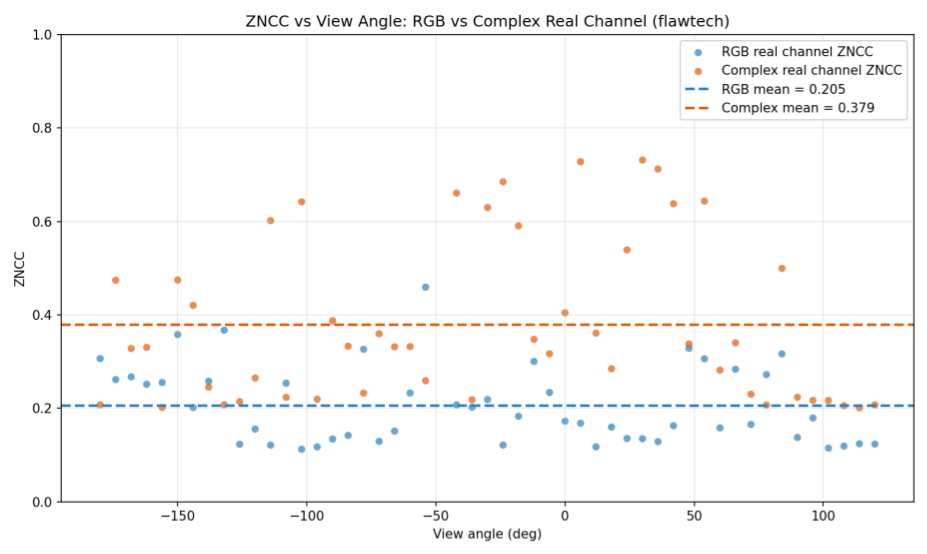}
  \caption{ZNCC relative to ground truth for the RGB and complex NeRF models at each scan angle for the T-shaped bracket.}\label{F_znccflaw}
\end{figure}

Complex-nerfacto achieved a higher and more stable correlation with ground truth for both test articles. The RGB nerfacto model produced a more erratic ZNCC curve with lower overall agreement (Figs.~\ref{F_zncccyl} and~\ref{F_znccflaw}). For both datasets, correlation dropped between $144^{\circ}$ and $168^{\circ}$, coinciding with a held-out angle in that region.

Both qualitatively and quantitatively (with ZNCC results), the complex-nerfacto showed more promise at preserving the complex-valued velocity data throughout the reconstruction pipeline. This yielded more accurate 3D surface wavefields when compared to representing the same data through conventional RGB images.

In early complex-nerfacto reconstructions, training on unmasked scan data rarely converged to a coherent 3D representation. We traced the cause of this behavior to laser speckle and calibration drift in the LDV's raster-scanned background, and small scanning artifacts and ``cloudiness'' present near the edges of the test article itself. These effects were not correlated to the rotation of the part, adjacent poses separated by $6^{\circ}$ showed substantial visual differences in the background region despite depicting very similar object views. This was difficult for the NeRF models to process, because these models rely on multi-view photometric consistency to generate correct color gradients. The network is assumed to observe a static scene from many angles, so consistent background noise was picked up as meaningful geometry in the NeRF renderings.

\begin{figure}[H]
  \centering
  \includegraphics[width=1\textwidth]{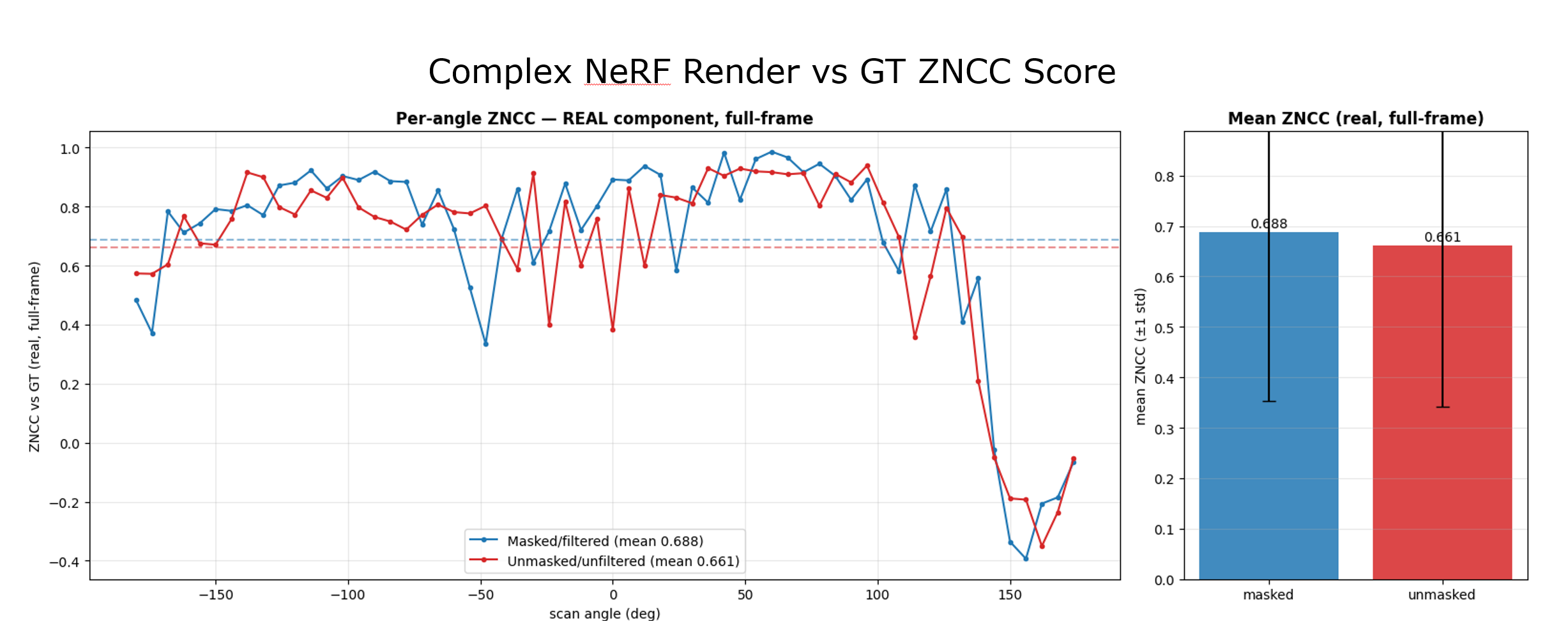}
  \caption{ZNCC alignment with ground truth: masked vs. unmasked training data for complex real NeRF.}
  \label{F_MVUZNCC}
\end{figure}

We addressed this issue with the rembg salient-object segmentation model to mask each scan, isolating the rotating test article and gripper from the surrounding background before training. We performed a controlled ablation study comparing NeRF models trained on masked and unmasked scans with camera intrinsics and training parameters. We found that masking substantially reduced the divergence and visual artifacts observed previously and produced a measurable increase in ZNCC score relative to ground truth (Fig.~\ref{F_MVUZNCC}). This result corroborated that background consistency was important for geometry abstraction, as expected for a NeRF model.

\section{Discussion and Conclusion}

This work demonstrates 3D reconstruction of an LDV surface-velocity wavefield using automated robotic manipulation and a measurement-preserving neural rendering pipeline. Robotic manipulation with embedded transducer is a promising method for collecting 3D wavefield measurements, and this method was used to generate the first renderings of a 3D surface velocity field. Our following contribution, complex-nerfacto, allows complex velocity matrices to be used directly as training data in the nerfacto pipeline, which avoids the information loss caused by quantizing small-magnitude signed velocity data into clamped RGB images.

As the results show, complex-nerfacto produces measurably higher fidelity and fewer artifacts than a baseline nerfacto pipeline trained on rendered LDV images. Because this is the first attempt to reconstruct an ASSESS wavefield in three dimensions, it establishes a working baseline while leaving substantial room for optimization.

The primary limitation of the current work is that sparse-input reconstruction was not achieved. High-fidelity renders required roughly 60 input scans, which is near-continuous angular coverage. At that scan density, the reconstruction's advantage over simply reviewing the 60 measured scans in sequence lies in the fact that the NeRF produces a single consistent 3D representation that can be queried at arbitrary viewing angles, rather than a set of independent 2D projections. Realizing the practical benefit of neural rendering, namely reduced acquisition time, requires reconstruction from far fewer views.

We see two possible routes to sparse-view reconstruction.

The first is architectural extension of nerfacto. Explicit part geometry could be estimated prior to training, either supplied directly as a geometric prior or enforced through a regularized variant such as RegNeRF or FreeNeRF \cite{bib17,bib24}. Because our masks already provide silhouettes at exactly known poses, a visual hull is available at no additional acquisition cost and could constrain the density field so that training capacity is spent on the wavefield rather than on recovering shape. A second candidate is the loss function. Both nerfacto and complex-nerfacto have output heads that are biased toward zero, which is also the LDV value assigned to masked background. This bias encourages collapse toward zero-valued regions, which is visible in the results. Treating the positive and negative velocity components separately, or weighting the loss by measured signal level so that low-carrier pixels contribute less, may mitigate this bias and better preserve high-magnitude wavefield features.

A second route is to use a different 3D representation entirely. Gaussian splatting \cite{bib25} is a promising candidate because its explicit primitives make it straightforward to attach a complex velocity value to a surface element. Alternatively, a LiDAR scanner co-located with the LDV beam could supply an independently measured point-cloud geometry to which velocity data are assigned directly, removing the need for the network to infer shape at all.

The broader motivation for reconstructing 3D wavefields is automated defect characterization. At present, defect identification still relies on a human inspector interpreting the 2D visualizations described in Section~\ref{S_viz}. A natural next step is to automate this stage using region-of-interest detection in the complex-gradient representation to identify cracks and delaminations. Combined with 3D reconstruction, this would enable an inspection workflow in which a robotic manipulator presents a part, the LDV captures its steady-state response, and a neural model reconstructs the surface wavefield and flags defects across the full geometry autonomously.

\section*{Acknowledgments}

This research was funded by Los Alamos National Laboratory (LANL) through the Engineering Institute's Los Alamos Dynamics Summer School. The Engineering Institute is a research and education collaboration between LANL and the University of California San Diego's Jacobs School of Engineering. This collaboration seeks to promote multidisciplinary engineering research that develops and integrates advanced predictive modeling, novel sensing systems, and new developments in information technology to address LANL mission-relevant problems.

The authors acknowledge Cole Maxwell, Erica Jacobson, and Scout Jarman for their support in this project.

\end{document}